\documentclass[sigconf]{acmart}

\def\BibTeX{{\rm B\kern-.05em{\sc i\kern-.025em b}\kern-.08emT\kern-.1667em\lower.7ex\hbox{E}\kern-.125emX}}

\setcopyright{none}

\usepackage{amssymb}
\usepackage{multirow}
\usepackage{graphicx}
\usepackage{enumerate}
\usepackage{subfigure}
\usepackage{bm}
\usepackage[ruled,linesnumbered]{algorithm2e}
\usepackage{amsmath}
\usepackage{float}
\usepackage{url}
\usepackage{hyperref}
\usepackage{enumitem}
\usepackage{threeparttable}
\usepackage{booktabs}
\usepackage[acronym,smallcaps,nowarn]{glossaries}
\usepackage{balance}
\usepackage{diagbox}

\copyrightyear{2019}
\acmYear{2019}

\allowdisplaybreaks[4]

\begin{document}

\title{Meta-GNN: On Few-shot Node Classification in Graph Meta-learning}

\author{Fan Zhou$^*$, Chengtai Cao$^*$, Kunpeng Zhang$^{\ddagger}$,  Goce Trajcevski$^{\mathsection}$, Ting Zhong$^*$, Ji Geng$^*$ }
\affiliation{%
	\institution{
	$^*$ University of Electronic Science and Technology of China, \\
	$^\ddagger$ University of Maryland, College park \\ $^{ \mathsection}$ Iowa State University, Ames 
	}
}

\begin{abstract}
Meta-learning has received a tremendous recent attention as a possible approach for mimicking human intelligence, i.e., acquiring new knowledge and skills with little or even no demonstration. Most of the existing meta-learning methods are proposed to tackle few-shot learning problems such as image and text, in rather Euclidean domain. However, there are very few works applying meta-learning to non-Euclidean domains, and the recently proposed graph neural networks (GNNs) models do not perform effectively on graph few-shot learning problems. Towards this, we propose a novel graph meta-learning framework -- Meta-GNN -- to tackle the few-shot node classification problem in graph meta-learning settings. It obtains the prior knowledge of classifiers by training on many similar few-shot learning tasks and then classifies the nodes from new classes with only few labeled samples. Additionally, Meta-GNN is a general model that can be straightforwardly incorporated into any existing state-of-the-art GNN. Our experiments conducted on three benchmark datasets demonstrate that our proposed approach not only improves the node classification performance by a large margin on few-shot learning problems in meta-learning paradigm, but also learns a more general and flexible model for task adaption.
\end{abstract}


\maketitle
\section{Introduction}
\label{Introduction}
Analyzing graph structure data with various deep learning methods has spurred a significant recent research interest~\cite{Wu2019}. A large number of models have been developed for solving different problems, including graph representation, link prediction, graph/node classification, etc. While earlier methods mainly focused on embedding nodes in an unsupervised manner in order to capture the global/local network structure, recent efforts applied sophisticated deep learning models -- which have been successfully used on Euclidean domain -- to non-Euclidean graph structure data, resulting in many prominent Graph Neural Networks (GNNs)-based methods -- e.g., GCN~\cite{kipf2016semi}, GraphSAGE~\cite{hamilton2017inductive}, SGC~\cite{sgc}, etc. 

Despite recent breakthroughs in GNNs, one setting that presents a persistent challenge is that of \textit{few-shot learning}, which aims at addressing data deficiency problem by recognizing new classes from very few labeled samples. Existing GNNs models always need to re-learn their parameters to incorporate the new information when new classes are encountered, and if the number of nodes in each new class is small, models' performance will suffer a catastrophic decline. The inability to handle situations where one or very few samples exist, is one of the major challenges for the current GNNs. To address this issue, some recent studies have focused on solving the few-shot learning on graph data where GNNs are either trained via co-training and self-training~\cite{li2018deeper}, or extended by stacking transposed graph convolutional layers for reconstructing input features~\cite{zhang2018few} -- both optimized for a single few-shot learning task with fixed classes.

Meta-learning (i.e., \textit{learning to learn}), has drawn a significant attention in the AI community in the recent years, due to its capability of quickly adapting to new tasks and learning transferable knowledge between tasks, with a few samples. It has been widely considered as similar to the human intelligence where humans are capable to rapidly learn new classes, with few samples and by utilizing previously learned prior knowledge. Meta-learning systems are trained by being exposed to a large number of tasks and are tested on their ability to learn new tasks. This differs from many standard machine learning techniques, which involve training on a single task and testing on held-out samples from that task~\cite{li2018deeper, zhang2018few}. We refer to the processes of learning on a large number of tasks and testing on a new task as \textit{meta-training} and \textit{meta-testing}, respectively. 

Adopting meta-learning for image and text learning has yielded significant progress and several models and algorithms have been recently proposed -- e.g., Matching Networks~\cite{vinyals2016matching}, Prototypical Networks~\cite{snell2017prototypical}, RelationNet~\cite{sung2018learning}, MAML~\cite{finn2017model}, etc. Despite the extensive studies and the promising results on analyzing data with Euclidean-like properties (e.g., images and text), there is surprisingly little work on applying meta-learning to graphs (non-Euclidean data). Among the main reasons are that graph data is more irregular, noisy and exhibits more complex relations among nodes, which make it difficult to directly apply existing meta-learning methods.

To bridge this gap, we present a general graph meta-learning framework called Meta-GNN, focusing on the few-shot node classification problem on graph data. To our knowledge, our work is the first to incorporate the meta-learning paradigm into GNNs, providing the capability of well adapting or generalizing to new classes that have never been encountered before, with very few samples. Instead of solely relying on the auxiliary information of nodes and aggregated information from neighbors in existing popular GNNs-based models, Meta-GNN is trained and optimized over numerous similar few-shot learning tasks towards better generalization of learning new tasks. In summary, the main contributions of our work are as follows:

\begin{itemize}[leftmargin=*]
\item We formulate a new graph few-shot learning paradigm for node classification. It is different from previous work in that we aim to classify nodes from new classes with only few samples each.
\item We propose a generic framework for tackling few-shot node classification, which can be easily combined with any popular GNNs models and opens up a new perspective of graph structure data analysis.
\item We demonstrate the superiority of our method over several state-of-the-art GNNs models on three benchmark datasets.
\end{itemize}

\section{Methodology}
\label{Methodology}
An undirected graph, denoted as $\mathcal{G}=(V, E, \mathbf{A}, \mathbf{X})$, is a quadruple where: (1) $V=\left\{v_{1}, v_{2}, \dots, v_{i}, \dots, v_{n}\right\}$ is a node set; (2) $E=\left\{e_{i, j} = (v_i, v_j) \right\} \subseteq (V \times V)$ is an edge set; (3) $\mathbf{A} \in \mathbb{R}^{n \times n}$ is a symmetric (typically sparse) adjacency matrix where $a_{i j}$ denotes the weight of the edge between nodes $v_{i}$ and $v_{j}$; and (4) $\mathbf{X} \in \mathbb{R}^{n \times d}$ is a feature matrix with $\mathbf{x}_{i} \in \mathbb{R}^{d}$ representing the characteristics of a given node $v_{i}$. 

\noindent\textbf{Problem definition}: 
We consider the few-shot node classification problem within the meta-learning paradigm. Simply speaking, we aim to obtain a classifier that can be adapted to new classes that are \textit{unseen} during the training process, given only a few samples in each new class. Formally, each node $v_i$ in the training set $\mathcal{D}_{\textit{train}}$ belongs to one of the classes in $C_{1}$. Nodes in a given disjoint testing set $\mathcal{D}_{\textit{test}}$ are associated with totally different new classes $C_{2}$. In $\mathcal{D}_{\textit{test}}$, a small number of nodes have labels/classes available. Our goal is to find a function $f_{\bm{\theta}}$ that is able to classify the rest of the unlabeled nodes into one of the classes in $C_{2}$, with a low misclassification rate. If the number of labeled nodes in each class is $K$, the task is known as $\left| C_{2} \right|$-way $K$-shot learning problem, where $K$ is a very small number. Thus, we address the graph meta-learning problem for node classification, in order to extract transferable knowledge from training data that will allow us to perform well on the testing data with unseen classes.

\subsection{Graph Neural Networks}
Modern GNNs jointly utilize the graph structure information and the node feature $\mathbf{X}$ to learn a new representation vector $\mathbf{h}_{v}$ of a node, usually following a neighborhood aggregation scheme. After $l$ iterations of the aggregation, a node's representation captures the structural information from its $l$-hop network neighbors. An illustrative example is shown in the right part of Figure~\ref{Frame}, whereby as the graph passes through the first layer of GNN, the red node aggregates the information from node 1, 2 and 3, and after the second layer, it gathers the information from node 5 and node 6. Formally, the $l$-th layer of a GNN is defined as:
\begin{align}
\label{AGGREGATE}
\mathbf{a}_{v}^{(l)}&={\mathbf{h}_{v}^{(l-1)}\cdot}\operatorname{AGGREGATE}^{(l)}\left(\left\{\mathbf{h}_{u}^{(l-1)} : u \in \mathcal{N}(v)\right\}\right), \\
\label{COMBINE}
\mathbf{h}_{v}^{(l)}&={\mathbf{h}_{v}^{(l-1)}\cdot}\operatorname{COMBINE}^{(l)}\left(\mathbf{a}_{v}^{(l)}\right),
\end{align}
where $\mathbf{h}_{v}^{(l)}$ is the feature vector of node $v$ at the $l$-iteration/layer. We initialize $\mathbf{h}^{(0)}=\mathbf{X}$ and denote with $\mathcal{N}(v)$ the set of nodes adjacent to $v$. The final representation $\mathbf{h}_{v}^{(L)}$ will be used for downstream tasks such as link prediction and node classification. The choices of AGGREGATE and COMBINE operation are crucial to the task performance, and plenty of approaches have been proposed and achieved impressive results (see~\cite{Wu2019} for a recent comprehensive overview). However, as observed in the literature~\cite{ravi2016optimization}, deep neural networks generally perform poorly on few-shot learning tasks. A main cause is that the standard gradient descent method (including its many variants) requires a large number of examples to obtain a satisfactory performance, which is not upheld in the few-shot learning setting.

\subsection{Meta-GNN}
We now present the details of our Meta-GNN framework. Our approach is based on the methodology of meta-learning which often follows an episodic paradigm and solves a new few-shot learning problem (i.e., meta-testing task, $\mathcal{T}_{mt}$) by training on many sampled similar tasks (i.e., meta-training tasks). We refer to the corresponding training and testing sets in all tasks as \textit{support set} and \textit{query set}, respectively. Our method leverages MAML~\cite{finn2017model} for the gradient updates during training. After training on considerable meta-training tasks, Meta-GNN is expected to learn (as a prior knowledge) how to quickly adapt to a new task using only a few datapoints in the new task. The performance of Meta-GNN is measured by meta-testing on the new task, i.e., fine-tuning Meta-GNN on a few samples from the support set of $\mathcal{T}_{mt}$ and evaluating it on the query set of $\mathcal{T}_{mt}$. 

We denote our Meta-GNN model as $f_{\bm{\theta}}$ with parameters ${\bm{\theta}}$, and the training set as $\mathcal{D}_{\textit{train}}=\left\{\left(\boldsymbol{x}_{1}, y_{1}\right), \ldots, \left(\boldsymbol{x}_{i}, y_{i}\right), \ldots, \left(\boldsymbol{x}_{N}, y_{N}\right)\right\}$ where $y_{i} \in  C_{1}$ and $N$ is the number of nodes in the training set. We will fork $M$ meta-training tasks from $\mathcal{D}_{\textit{train}}$: ${\mathcal{T}}=\{\mathcal{T}_{1},\mathcal{T}_{2},\cdots,\mathcal{T}_{M}\}$. Both the support set $\mathcal{S}_{i}$ and query set $\mathcal{Q}_{i}$ of meta-training task $\mathcal{T}_{i}$ are sampled from $\mathcal{D}_{\textit{train}}$. In the support set, we have $\mathcal{S}_{i}=\left\{v_{i1}, v_{i2}, \dots, v_{is}\right\}=\left\{\left(\boldsymbol{x}_{i1}, y_{i1}\right),\left(\boldsymbol{x}_{i2}, y_{i2}\right), \ldots,\left(\boldsymbol{x}_{is}, y_{is}\right)\right\}$, where $s = |\mathcal{S}_{i}|$; $\boldsymbol{x}_{is}$ is the input vector of node ${v_{is}}$ with label $y_{is}$.

\begin{figure}[!ht]\small
	\centering
	\includegraphics[width=0.35\textwidth]{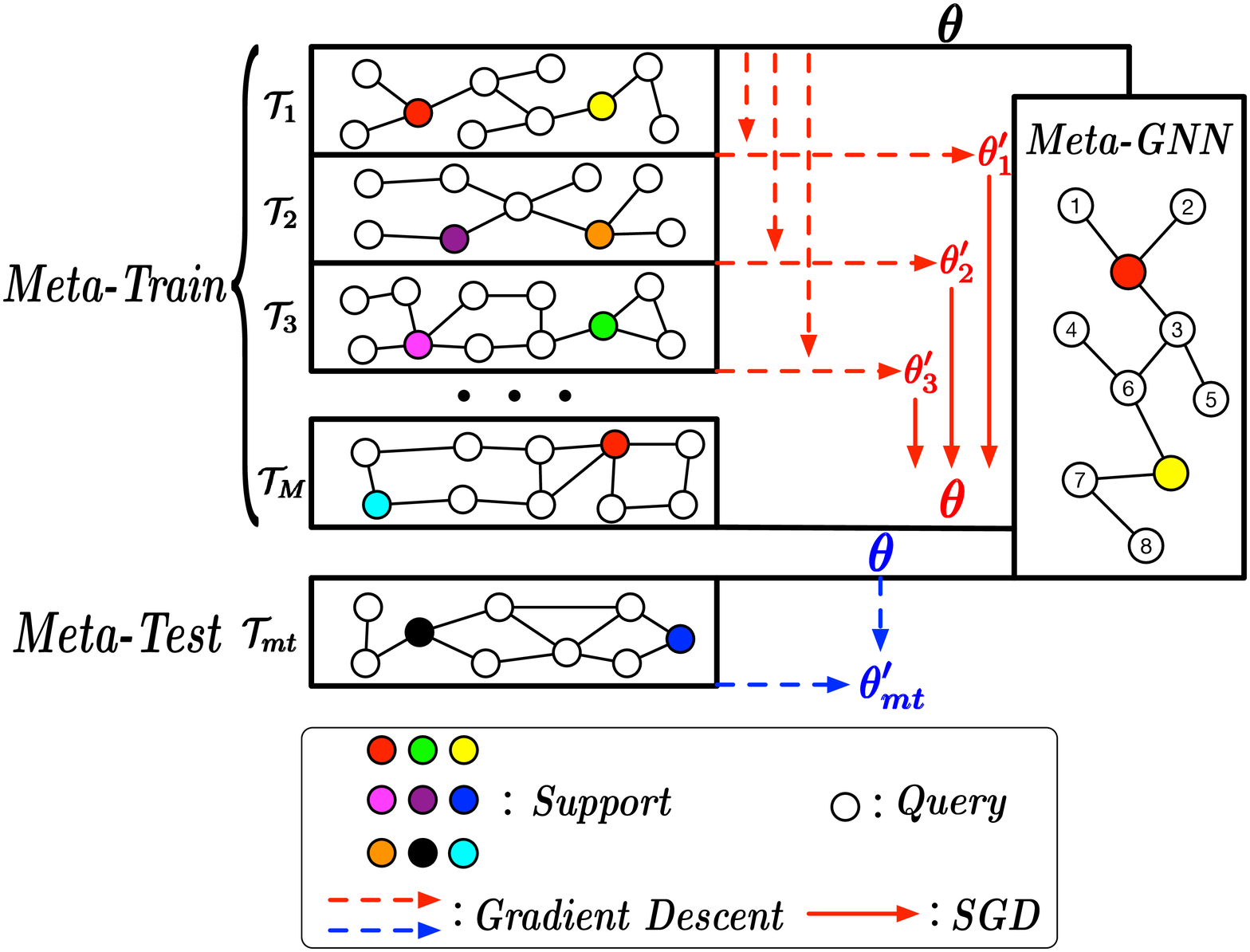}
	\caption{Overview of Meta-GNN. The black $\bm{\theta}$, red $\bm{\theta}$ and blue $\bm{\theta}$ represent the randomly initialized parameters, the parameters after one meta-update and after all meta-updates, respectively. $\bm{M}$ is the number of meta-training tasks.}
	\label{Frame}
\end{figure}

\noindent
\textbf{Task Sampling}:
\label{Task_sampling}Before presenting the details of Meta-GNN, we prepare $M$ tasks $\mathcal{T}$ by sampling from $\mathcal{D}_{\textit{train}}$ according to the graph meta-learning principals. In short, we sample $\left| C_{2}\right|$ classes from $C_{1}$ and then randomly sample $K$ nodes for each class to simulate few-shot node classification. Specifically, we use the following steps to generate the graph meta-training tasks:
\begin{enumerate}
\item $C\leftarrow$ RANDOMSAMPLE $\left( C_{1}, \left| C_{2}\right| \right)$;
\item $\mathcal{S}_{i}\leftarrow$ RANDOMSAMPLE $\left(\mathcal{D}_{C}, K \times \left| C_{2}\right| \right)$;
\item $\mathcal{Q}_{i}\leftarrow$ RANDOMSAMPLE $\left(\mathcal{D}_{C}-\mathcal{S}_{i}, P\right)$;
\item $\mathcal{T}_{i} = \mathcal{S}_{i} + \mathcal{Q}_{i}$;
\item Repeat step (1) - (4) for $M$ times;
\end{enumerate}
Thus, we first randomly sample $\left| C_{2}\right|$ classes from $C_{1}$, denoted as $C$. Then we can obtain $\mathcal{D}_{C}$, which is a subset of training set $\mathcal{D}_{\textit{train}}$ containing elements $\left(\boldsymbol{x}_{i}, y_{i}\right)$, where $y_{i}$ is one of the classes in $C$; Next, we randomly sample $K \times \left| C_{2}\right| $ nodes from $\mathcal{D}_{C}$ to form the support set $\mathcal{S}_{i}$, where $K$ is the number of nodes in each class of $\mathcal{S}_{i}$ (i.e., the number of shots). Finally, we randomly sample $P$ nodes from the remaining nodes in $\mathcal{D}_{C}$ to compose the query set $\mathcal{Q}_{i}$ which, in turn, is used in constructing the meta-training task $\mathcal{T}_{i} = \mathcal{S}_{i} + \mathcal{Q}_{i}$. Repeating the above steps $M$ times, yields $M$ meta-training tasks.

\noindent
\textbf{Meta-training}:
\label{Meta_training}We expect to obtain a good initialization of Meta-GNN, which is generally applicable to similar tasks, and explicitly encourage the initialization parameters to perform well after a small number of gradient descent updates on a new few-shot learning task. When learning a task $\mathcal{T}_{i}$, we begin with feeding the support set $\mathcal{S}_{i}$ to Meta-GNN, and calculate the cross-entropy loss:
\begin{small}
\begin{align}
\label{Loss}
\mathcal{L}_{\mathcal{T}_{i}}\left(f_{\bm{\theta}}\right) = - \left(\sum_{\boldsymbol{x}_{is}, y_{is}} y_{is} \log f_{\bm{\theta}}\left(\boldsymbol{x}_{is}\right) + \left(1-y_{is}\right) \log \left(1-f_{\bm{\theta}}\left(\boldsymbol{x}_{is}\right)\right) \right).
\end{align}\end{small}

Then we perform parameters updates, using a simple gradient descent with one or several steps in task $\mathcal{T}_{i}$. For brevity, we will only describe one gradient update in the rest of this section -- with a note that performing multiple gradient updates is a straightforward extension:
\begin{equation}
\label{Gradient_Descent}
\bm{{\theta}_{i}^{\prime}}=\bm{\theta}-\alpha_{1}\frac{\partial \mathcal{L}_{\mathcal{T}_{i}}\left(f_{\bm{\theta}}\right)}{\partial \bm{\theta}}
\end{equation}
where $\alpha_{1}$ is the task-learning rate and the model parameters are trained to optimize the performance of $f_{\bm{{\theta}_{i}^{\prime}}}$ across meta-training tasks. More specifically, the meta-objective is as follows:
\begin{equation}
\label{Meta_Objective}
\bm{\theta} = \mathop{\arg\min}_{\bm{\theta}} \sum_{\mathcal{T}_{i} \sim p(\mathcal{T})} \mathcal{L}_{\mathcal{T}_{i}}\left(f_{\bm{{\theta}_{i}^{\prime}}}\right)
\end{equation}
where $p(\mathcal{T})$ is the distribution of meta-training tasks. Note that the meta-optimization is performed over the model parameters $\bm{\theta}$, whereas the objective is computed using the updated model parameters $\bm{{\theta}^{\prime}}$. This is because we need good initialization parameters $\bm{\theta}$ for all the similar few-shot node classification tasks, instead of certain updated parameters $\bm{{\theta}^{\prime}_{i}}$ performing well on only a particular task $\mathcal{T}_{i}$. Essentially, Meta-GNN aims to optimize the model parameters so that it maximizes the node classification performance on a new task after one or a small number of gradient descent updates.
The meta-optimization across tasks is performed via stochastic gradient descent (SGD), and the model parameters $\bm{\theta}$ are updated as follows:
\begin{equation}
\label{Meta_Update}
\bm{\theta} \leftarrow \bm{\theta}-\alpha_{2}\frac{\partial \sum_{\mathcal{T}_{i} \sim p(\mathcal{T})} \mathcal{L}_{\mathcal{T}_{i}}\left(f_{\bm{{\theta}_{i}^{\prime}}}\right)}{\partial \bm{\theta}}
\end{equation}
where $\alpha_{2}$ is the meta-learning rate -- i.e., an additional hyperparameter introduced by our framework.

\noindent
\textbf{Meta-testing}:
For  meta-testing, we simply need to feed the nodes of the support set of the new few-shot learning task (i.e., $\mathcal{T}_{mt}$) to the Meta-GNN, and update parameters $\bm{{\theta}^{\prime}_{mt}}$ via one or a small number of gradient descent steps using Eq.(\ref{Gradient_Descent}). Therefore, the performance of Meta-GNN can be easily evaluated on the query set of $\mathcal{T}_{mt}$. 

The meta-training and meta-testing procedures of Meta-GNN are outlined in Algorithm~\ref{alg:algorithm1} and the overall operational process of Meta-GNN is depicted in Figure~\ref{Frame}.

\begin{algorithm}[t]
	\caption{Framework of Meta-GNN.}
	\small
	\label{alg:algorithm1}
	\SetAlgoLined
	\DontPrintSemicolon
	\BlankLine
	\KwIn{Distribution over mete-training tasks: $p(\mathcal{T})$; Meta-testing tasks: $\mathcal{T}_{mt}$; Task-learning rate: $\alpha_{1}$; Meta-learning rate: $\alpha_{2}$.}
	\KwOut{Labels of nodes in query set of $\mathcal{T}_{mt}$.}  
	\BlankLine
	Initialize $\bm{\theta}$ randomly;
	
	\While{\textnormal{not converged}}{
		Sample batch of meta-training tasks $\mathcal{T}_{i} \sim p(\mathcal{T})$;
		
		\ForEach{task in $\mathcal{T}_{i}$}{
			Evaluate $\mathcal{L}_{\mathcal{T}_{i}}\left(f_{\bm{\theta}}\right)$ using $\mathcal{S}_{i}$ by Eq.~(\ref{Loss});
			
			Compute adapted parameters $\bm{{\theta}^{\prime}_{i}}$ by Eq.~(\ref{Gradient_Descent});
			
			Evaluate $\mathcal{L}_{\mathcal{T}_{i}}\left(f_{\bm{{\theta}_{i}^{\prime}}}\right)$ using $\mathcal{Q}_{i}$ by Eq.~(\ref{Loss});
		}
		Update $\bm{\theta}$ by Eq.~(\ref{Meta_Update}); 
	}
	
	Compute adapted parameters $\bm{{\theta}^{\prime}_{mt}}$ using support set of $\mathcal{T}_{mt}$ by Eq.~(\ref{Gradient_Descent}); 
	
	Predict labels of nodes in query set of $\mathcal{T}_{mt}$ using model $f_{\bm{{\theta}_{mt}^{\prime}}}$.

\end{algorithm}

\section{Experiments}
\label{Experiments}
We now present the empirical results of Meta-GNN performance for node classification on three commonly used datasets: Cora~\cite{sen2008collective}, Citeseer~\cite{sen2008collective} and Reddit~\cite{hamilton2017inductive}. For reproducibility, the source code and datasets are publicly available\footnote{\url{https://github.com/AI-DL-Conference/Meta-GNN}}. Since our work focuses on few-shot learning problem in the context of meta-learning paradigm, we make some modifications in the dataset partition, to ensure the feasibility of the few-shot learning (cf. Section~\ref{Task_sampling}). 

\begin{table}[!ht]
	\caption{Descriptive statistics and partition of three datasets.}
	\label{Datasets} 
	\begin{tabular}{r|ccc}
		\hline
		&Cora &Citeseer &Reddit \\
		\hline
		\text{\#} Nodes 	&2,708 &3,327 &232,965\\
		\text{\#} Features 	&1,433 &3,703 &602\\
		\text{\#} Labels 	&7 &6 &41\\
		\hline
		\text{\#} $\left| C_{1} \right|$	&5 &4 &36\\
		\text{\#} $\left| C_{2} \right|$	&2 &2 &5\\
		\hline
	\end{tabular}
\end{table}
For the Cora and Citeseer datasets, we randomly set apart two classes as meta-testing classes (i.e., the corresponding nodes forming $\mathcal{D}_{\textit{test}}$ with $ \left| C_{2} \right| = 2 $), and the rest nodes (i.e., $\mathcal{D}_{\textit{train}}$ ) are used to generate meta-training tasks following the method described in section~\ref{Task_sampling}. Similarly for the Reddit, except that $\left| C_{2} \right| = 5$ due to the relatively large number of unique labels. Each class in the support set of meta-training and meta-testing tasks has only one or three samples (i.e., $K = 1$ or $K = 3$) for all datasets. We observed that, when the support set is extremely small, the performance of each method is sensitive to the node selection. Therefore, we evaluate all models on 50 randomly selected nodes in the support set and report the average accuracy on Cora and Citeseer. For Reddit, as there are significant more classes and corresponding nodes in each class, we select the same nodes as support set for each run and conduct 50 different runs, and report the average Micro-F1 score. In order to fairly and accurately assess performance of our Meta-GNN and other baselines, we conduct 10 cross-validation on Cora and Reddit and 5 on Citeseer. The statistics of datasets after pre-processing are shown in Table~\ref{Datasets}.

\noindent
\textbf{Baselines}:
For Cora and Citeseer, we compare Meta-GNN against DeepWalk~\cite{perozzi2014deepwalk}, Node2Vec~\cite{grover2016node2vec}, GCN~\cite{kipf2016semi}, SGC~\cite{sgc} and GraphSAGE~\cite{hamilton2017inductive}. For Reddit, we compare with GCN~\cite{kipf2016semi}, SGC~\cite{sgc} and GraphSAGE~\cite{hamilton2017inductive} since our implementation on other baselines experienced memory explosion or unbearable training time. Notably, we only modify the dataset partition to satisfy the few-shot learning setting in meta-learning paradigm, and other settings of each model are the same as its original implementation.

\noindent
\textbf{Implementation and Parameter Setups}:
To demonstrate the applicability of our framework, we implement Meta-GNN with two GNN models, i.e., SGC and GCN, forming two instances Meta-SGC and Meta-GCN. To accelerate the model convergence, we set batch size (line 3 in Algorithm ~\ref{alg:algorithm1}) as 5 for Cora and Citeseer, but 12 for Reddit. The $\alpha_{1}$ and $\alpha_{2}$ are set to 0.5 and 0.003 in Meta-SGC, while the corresponding values are set to 0.1 and 0.001 in Meta-GCN respectively. The other settings of GNN module in the two models are the same as the suggestion of their original papers.

\noindent
\textbf{Experimental Result}:
Table~\ref{Cora_Citeseer_Results} and Table~\ref{Reddit_Result} show the results of the performance comparison between Meta-GNN and baselines, from which we can clearly observe that our proposed model achieves the best performance across all three datasets. Generally speaking, GNN models, including ours, significantly outperform the embedding based models, e.g., DeepWalk and Node2Vec, on the few-shot learning scenarios. Among GNN models, we surprisingly find that two inductive GNN models, GraphSAGE with Mean and Pool operation/variants, have not shown competitive results, even compared with GCN and SGC. This result indicates that previous inductive graph learning models, which have shown promising results when encountering with \textit{new nodes} (as reported in the original paper~\cite{hamilton2017inductive}), do not generalize well to \textit{new classes}.

The superiority of Meta-GNN on Cora and Citeseer varies with the number of samples in support set, i.e., the less samples, the more improvement Meta-GNN improves over the baselines. This proves our primary motivation, i.e., adapting meta-learning into GNN models for few-shot graph learning. On the more challenging dataset Reddit, Meta-GNN can reap more improvement due to its capability of adapting to new tasks -- there are more tasks for learning to learn the node representation and therefore a more general task adaption model obtained by Meta-GNN.

When comparing the two GNN modules in Meta-GNN, we did not observe any significant discrepancy -- Meta-SGC performs better on Cora while Meta-GCN achieve slightly higher scores on Citeseer and Reddit, which implies that existing GCN based models do not beat each other on node classification performance -- however we note that SGC and its corresponding meta-learning version Meta-SGC are ordered faster than GCN and Meta-GCN~\cite{sgc}, respectively.

\begin{table}[!ht]
	\centering
	\renewcommand{\multirowsetup}{
		\centering}
	\caption{Accuracy comparison: Meta-GNN vs. baselines. }
	\label{Cora_Citeseer_Results}
	\small 
	\begin{tabular}{@{}c|m{4em}<{\centering}|m{4em}<{\centering}|m{4em}<{\centering}|m{4em}<{\centering}}
		\hline
		Datasets& \multicolumn{2}{c|}{Cora} & \multicolumn{2}{c}{Citeseer}\\
		\hline
		\diagbox[width=8em,trim=l]{Model}{$K$-shot} &1-shot &3-shot &1-shot &3-shot \\
		\hline
		DeepWalk & 16.06\% & 25.67\% & 14.52\% & 21.18\%\\
		\hline
		Node2Vec & 15.15\% & 25.66\% & 12.98\% & 20.02\%\\
		\hline
		GraphSAGE-Mean & 50.89\% & 53.12\%. & 53.49\% & 55.01\%\\
		\hline
		GraphSAGE-Pool & 48.53\% & 50.15\% & 51.02\% & 53.98\%\\
		\hline
		SGC 	 & 61.64\% & 75.67\% & 56.91\% & 65.67\%\\
		\hline
		GCN 	 & 60.33\% & 75.15\% & 58.44\% & 67.99\%\\
		\hline
		Meta-SGC & \textbf{65.27\%}& \textbf{77.19\%} & 60.46\%  & 68.65\% \\
		\hline
		Meta-GCN & 63.72\% & 76.78\% & \textbf{61.91\%} & \textbf{69.43\%} \\
		\hline
	\end{tabular}
\end{table}

\begin{table}[!ht]
	\centering
	\renewcommand{\multirowsetup}{
		\centering}
	\caption{Micro-F1 comparison: Meta-GNN vs. baselines. }
	\label{Reddit_Result}
	\small 
	\begin{tabular}{@{}c|m{5em}<{\centering}|m{5em}<{\centering}}
		\hline
		Datasets& \multicolumn{2}{c}{Reddit} \\
		\hline
		\diagbox[width=10em,trim=l]{Model}{$K$-shot} &1-shot &3-shot\\
		\hline
		GraphSAGE-Mean & 9.47\% & 15.89\% \\
		\hline
		GraphSAGE-Pool & 9.31\% & 15.36\% \\
		\hline
		SGC 	 & 9.80\% & 16.98\% \\
		\hline
		GCN 	 & 9.87\% & 17.17\% \\
		\hline
		Meta-SGC & 14.15\% & 20.14\%\\
		\hline
		Meta-GCN & \textbf{14.22}\% & \textbf{20.71}\%\\
		\hline
	\end{tabular}
\end{table}

\section{Conclusions}
\label{Conclusions}
We have presented a generic graph meta-learning framework for few-shot node classification that leverages meta-learning mechanism to learn better parameter initialization of GNNs. The proposed Meta-GNN model can adapt well to new learning tasks (even new classes) with few labeled samples and significantly improves the performance in the context of few-shot node classification under meta-learning paradigm. Encouraging results have been obtained on three widely used datasets. In our future work, we would like to extend our framework to address more challenging problems such as few-shot graph classification and zero-shot node classification.

\bibliographystyle{ACM-Reference-Format}
\bibliography{MetaGNN}


\begin{thebibliography}{14}


\ifx \showCODEN    \undefined \def \showCODEN     #1{\unskip}     \fi
\ifx \showDOI      \undefined \def \showDOI       #1{#1}\fi
\ifx \showISBNx    \undefined \def \showISBNx     #1{\unskip}     \fi
\ifx \showISBNxiii \undefined \def \showISBNxiii  #1{\unskip}     \fi
\ifx \showISSN     \undefined \def \showISSN      #1{\unskip}     \fi
\ifx \showLCCN     \undefined \def \showLCCN      #1{\unskip}     \fi
\ifx \shownote     \undefined \def \shownote      #1{#1}          \fi
\ifx \showarticletitle \undefined \def \showarticletitle #1{#1}   \fi
\ifx \showURL      \undefined \def \showURL       {\relax}        \fi
\providecommand\bibfield[2]{#2}
\providecommand\bibinfo[2]{#2}
\providecommand\natexlab[1]{#1}
\providecommand\showeprint[2][]{arXiv:#2}

\bibitem[\protect\citeauthoryear{Finn, Abbeel, and Levine}{Finn
  et~al\mbox{.}}{2017}]%
        {finn2017model}
\bibfield{author}{\bibinfo{person}{Chelsea Finn}, \bibinfo{person}{Pieter
  Abbeel}, {and} \bibinfo{person}{Sergey Levine}.}
  \bibinfo{year}{2017}\natexlab{}.
\newblock \showarticletitle{Model-agnostic meta-learning for fast adaptation of
  deep networks}. In \bibinfo{booktitle}{\emph{ICML}}.
\newblock


\bibitem[\protect\citeauthoryear{Grover and Leskovec}{Grover and
  Leskovec}{2016}]%
        {grover2016node2vec}
\bibfield{author}{\bibinfo{person}{Aditya Grover} {and} \bibinfo{person}{Jure
  Leskovec}.} \bibinfo{year}{2016}\natexlab{}.
\newblock \showarticletitle{node2vec: Scalable feature learning for networks}.
  In \bibinfo{booktitle}{\emph{KDD}}.
\newblock


\bibitem[\protect\citeauthoryear{Hamilton, Ying, and Leskovec}{Hamilton
  et~al\mbox{.}}{2017}]%
        {hamilton2017inductive}
\bibfield{author}{\bibinfo{person}{Will Hamilton}, \bibinfo{person}{Zhitao
  Ying}, {and} \bibinfo{person}{Jure Leskovec}.}
  \bibinfo{year}{2017}\natexlab{}.
\newblock \showarticletitle{Inductive representation learning on large graphs}.
  In \bibinfo{booktitle}{\emph{NIPS}}.
\newblock


\bibitem[\protect\citeauthoryear{Kipf and Welling}{Kipf and Welling}{2017}]%
        {kipf2016semi}
\bibfield{author}{\bibinfo{person}{Thomas~N Kipf} {and} \bibinfo{person}{Max
  Welling}.} \bibinfo{year}{2017}\natexlab{}.
\newblock \showarticletitle{Semi-supervised classification with graph
  convolutional networks}. In \bibinfo{booktitle}{\emph{ICLR}}.
\newblock


\bibitem[\protect\citeauthoryear{Li, Han, and Wu}{Li et~al\mbox{.}}{2018}]%
        {li2018deeper}
\bibfield{author}{\bibinfo{person}{Qimai Li}, \bibinfo{person}{Zhichao Han},
  {and} \bibinfo{person}{Xiao-Ming Wu}.} \bibinfo{year}{2018}\natexlab{}.
\newblock \showarticletitle{Deeper insights into graph convolutional networks
  for semi-supervised learning}. In \bibinfo{booktitle}{\emph{AAAI}}.
\newblock


\bibitem[\protect\citeauthoryear{Perozzi, Al-Rfou, and Skiena}{Perozzi
  et~al\mbox{.}}{2014}]%
        {perozzi2014deepwalk}
\bibfield{author}{\bibinfo{person}{Bryan Perozzi}, \bibinfo{person}{Rami
  Al-Rfou}, {and} \bibinfo{person}{Steven Skiena}.}
  \bibinfo{year}{2014}\natexlab{}.
\newblock \showarticletitle{Deepwalk: Online learning of social
  representations}. In \bibinfo{booktitle}{\emph{KDD}}.
\newblock


\bibitem[\protect\citeauthoryear{Ravi and Larochelle}{Ravi and
  Larochelle}{2016}]%
        {ravi2016optimization}
\bibfield{author}{\bibinfo{person}{Sachin Ravi} {and} \bibinfo{person}{Hugo
  Larochelle}.} \bibinfo{year}{2016}\natexlab{}.
\newblock \showarticletitle{Optimization as a model for few-shot learning}. In
  \bibinfo{booktitle}{\emph{ICLR}}.
\newblock


\bibitem[\protect\citeauthoryear{Sen, Namata, Bilgic, Getoor, Galligher, and
  Eliassi-Rad}{Sen et~al\mbox{.}}{2008}]%
        {sen2008collective}
\bibfield{author}{\bibinfo{person}{Prithviraj Sen}, \bibinfo{person}{Galileo
  Namata}, \bibinfo{person}{Mustafa Bilgic}, \bibinfo{person}{Lise Getoor},
  \bibinfo{person}{Brian Galligher}, {and} \bibinfo{person}{Tina Eliassi-Rad}.}
  \bibinfo{year}{2008}\natexlab{}.
\newblock \showarticletitle{Collective classification in network data}.
\newblock \bibinfo{journal}{\emph{AI magazine}} \bibinfo{volume}{29},
  \bibinfo{number}{3} (\bibinfo{year}{2008}), \bibinfo{pages}{93--93}.
\newblock


\bibitem[\protect\citeauthoryear{Snell, Swersky, and Zemel}{Snell
  et~al\mbox{.}}{2017}]%
        {snell2017prototypical}
\bibfield{author}{\bibinfo{person}{Jake Snell}, \bibinfo{person}{Kevin
  Swersky}, {and} \bibinfo{person}{Richard Zemel}.}
  \bibinfo{year}{2017}\natexlab{}.
\newblock \showarticletitle{Prototypical networks for few-shot learning}. In
  \bibinfo{booktitle}{\emph{NIPS}}.
\newblock


\bibitem[\protect\citeauthoryear{Sung, Yang, Zhang, Xiang, Torr, and
  Hospedales}{Sung et~al\mbox{.}}{2018}]%
        {sung2018learning}
\bibfield{author}{\bibinfo{person}{Flood Sung}, \bibinfo{person}{Yongxin Yang},
  \bibinfo{person}{Li Zhang}, \bibinfo{person}{Tao Xiang},
  \bibinfo{person}{Philip~HS Torr}, {and} \bibinfo{person}{Timothy~M
  Hospedales}.} \bibinfo{year}{2018}\natexlab{}.
\newblock \showarticletitle{Learning to compare: Relation network for few-shot
  learning}. In \bibinfo{booktitle}{\emph{CVPR}}.
\newblock


\bibitem[\protect\citeauthoryear{Vinyals, Blundell, Lillicrap, Wierstra,
  et~al\mbox{.}}{Vinyals et~al\mbox{.}}{2016}]%
        {vinyals2016matching}
\bibfield{author}{\bibinfo{person}{Oriol Vinyals}, \bibinfo{person}{Charles
  Blundell}, \bibinfo{person}{Timothy Lillicrap}, \bibinfo{person}{Daan
  Wierstra}, {et~al\mbox{.}}} \bibinfo{year}{2016}\natexlab{}.
\newblock \showarticletitle{Matching networks for one shot learning}. In
  \bibinfo{booktitle}{\emph{NIPS}}.
\newblock


\bibitem[\protect\citeauthoryear{Wu, Zhang, Souza~Jr., Fifty, Yu, and
  Weinberger}{Wu et~al\mbox{.}}{2019b}]%
        {sgc}
\bibfield{author}{\bibinfo{person}{Felix Wu}, \bibinfo{person}{Tianyi Zhang},
  \bibinfo{person}{Amauri~Holanda Souza~Jr.}, \bibinfo{person}{Christopher
  Fifty}, \bibinfo{person}{Tao Yu}, {and} \bibinfo{person}{Kilian~Q.
  Weinberger}.} \bibinfo{year}{2019}\natexlab{b}.
\newblock \showarticletitle{Simplifying Graph Convolutional Networks}. In
  \bibinfo{booktitle}{\emph{ICML}}.
\newblock


\bibitem[\protect\citeauthoryear{Wu, Pan, Chen, Long, Zhang, and Yu}{Wu
  et~al\mbox{.}}{2019a}]%
        {Wu2019}
\bibfield{author}{\bibinfo{person}{Zonghan Wu}, \bibinfo{person}{Shirui Pan},
  \bibinfo{person}{Fengwen Chen}, \bibinfo{person}{Guodong Long},
  \bibinfo{person}{Chengqi Zhang}, {and} \bibinfo{person}{Philip~S Yu}.}
  \bibinfo{year}{2019}\natexlab{a}.
\newblock \showarticletitle{A Comprehensive Survey on Graph Neural Networks}.
\newblock \bibinfo{journal}{\emph{arXiv.org}} (\bibinfo{year}{2019}).
\newblock


\bibitem[\protect\citeauthoryear{Zhang, Zhou, Huang, and Wei}{Zhang
  et~al\mbox{.}}{2018}]%
        {zhang2018few}
\bibfield{author}{\bibinfo{person}{Shengzhong Zhang}, \bibinfo{person}{Ziang
  Zhou}, \bibinfo{person}{Zengfeng Huang}, {and} \bibinfo{person}{Zhongyu
  Wei}.} \bibinfo{year}{2018}\natexlab{}.
\newblock \showarticletitle{Few-shot Classification on Graphs with Structural
  Regularized GCNs}.
\newblock \bibinfo{journal}{\emph{arXiv preprint}} (\bibinfo{year}{2018}).
\newblock


\end{thebibliography}

\end{document}